\documentclass[a4paper]{article}

\usepackage{INTERSPEECH2015}

\usepackage{amssymb,bm}
\usepackage{textcomp}

\usepackage{amsmath,dsfont,graphicx}
\usepackage{mathrsfs}
\usepackage{epstopdf}

\newcommand{\rpm}{\sbox0{$1$}\sbox2{$\scriptstyle\pm$}
  \raise\dimexpr(\ht0-\ht2)/2\relax\box2 }

\DeclareMathOperator{\tr}{tr}

\ninept

\title{Blind score normalization method for PLDA based speaker recognition}

\makeatletter
\def\name#1{\gdef\@name{#1\\}}
\makeatother
\name{{\em Danila Doroshin$^1$, Nikolay Lubimov$^1$,}
      {\em Marina Nastasenko$^1$, Mikhail Kotov$^1$}}
\address{$^1$Stel - Computer Systems Ltd., Moscow, Russia \\
  {\small \tt \{doroshin, lubimov, marina.nastasenko, kotov\}@stel.ru}
}

\begin{document}

\maketitle
\begin{abstract}
Probabilistic Linear Discriminant Analysis (PLDA) has become state-of-the-art method for modeling $i$-vector space in speaker recognition task. However the performance degradation is observed if enrollment data size differs from one speaker to another. This paper presents a solution to such problem by introducing new PLDA scoring normalization technique. Normalization parameters are derived in a blind way, so that, unlike traditional \textit{ZT-norm}, no extra development data is required. Moreover, proposed method has shown to be optimal in terms of detection cost function. The experiments conducted on NIST SRE 2014 database demonstrate an improved accuracy in a mixed enrollment number condition.
\end{abstract}

\section{Introduction}

One trial of automatic speaker verification process consists of estimating probability whether two utterances belong to the same speaker. Typically one utterance is taken from enrollment set of target speakers, another one is unknown test utterance. 
The decision is taken by the comparison of the verification score with a threshold. The combination of $i$-vector and Probabilistic Linear Discriminant Analysis (PLDA) approaches is a state-of-the-art speaker recognition method allowing to obtain this score. The $i$-vector approach is based on the total variability model \cite{dehak2011front} representing speaker data in the low-dimensional space.
PLDA \cite{prince2007probabilistic}  handles the influence of the channel variability in the $i$-vector space \cite{kenny2008study} and also enables to compute the Log-Likelihood Ratio (LLR) score between \textit{target} and \textit{non-target} hypothesis. There are two concurrent hypothesis forming this LLR score based on dependency of random latent variables which make probabilistic inference of visible data.

There are various strategies for PLDA score obtaining depending on the enrollment data size \cite{rajan2014single}. The approach with multiple enrollment utterances shows the best results among other strategies in the text-dependent verification
\cite{larcher2013phonetically}. However in real applications it often happens that the available data is limited and varied from one speaker to another. Using classical PLDA modeling approach in this case, system performance may be unstable depending on the enrollment size. In this paper, we introduce new PLDA scoring normalization technique that deals with this variation by minimizing detection cost function. Unlike traditional \textit{ZT-norm} that is used to handle the effect of new environment \cite{barras2003feature}, the introduced method handles the effect of different quality of speaker models and does not require extra development data. Speaker dependent score distributions for  \textit{target} and \textit{non-target} hypothesis are considered  to pose the problem. 

The paper is organized as follows. In section \ref{sec:optthresh}, the proposed PLDA scores normalization technique is presented. In section \ref{sec:scorenorm}, the process of estimation of speaker dependent distribution parameters is presented.
In section \ref{sec:experiments} the train and test data sets are described. The results for NIST SRE 2014 data set are given. In section \ref{sec:conclusions}, conclusions and future work directions are discussed.
\section{Score normalization for normal distributed scores}
\label{sec:optthresh}
Consider \textit{speaker dependent} Gaussian distributions for PLDA LLR scores and corresponding Cumulative Distribution Functions (CDF) in the cases of target $H_1$ and non-target hypothesis $H_2$
\begin{eqnarray}
\label{init_tar}
P_r(s | H_1) = \mathcal{N}(s, \mu_1, \sigma_1^2), \quad \Phi_{S}(t | H_1) = P_r(S < t| H_1)
\\
\label{init_nontar}
P_r(s | H_2) = \mathcal{N}(s, \mu_2, \sigma_2^2), \quad \Phi_{S}(t | H_2) = P_r(S < t| H_2)
\end{eqnarray}
where $r$ is an index of the speaker, $\mu_1, \sigma_1, \mu_2, \sigma_2$ -- speaker dependent parameters, $s$ is a PLDA LLR score. 
The choice of Gaussian distribution for scores will be discussed in section \ref{sec:scorenorm}.
We use minDCF as a measure of the system performance: $minDCF = \min_{t} ({FR}(t) + \beta {FA}(t))$, where $FA$ and $FR$ denote the false acceptance and the false rejection rates, $\beta$ is a fixed constant, and $t$ the varying threshold.

It is expected that the value of the speaker specific threshold depends on the quality of the speaker model. This quality depends on the speaker's enrollment size, to what extent this enrollment represents speakers speech and also depends on the environment during enrollment recording.

The process of deriving the optimal threshold that is individual for each speaker is considered below.  Rewrite minDCF in terms of CDF
\begin{equation}
\label{mindcf}
minDCF = \min_{t} \Phi_S(t | H_1) + \beta (1 - \Phi_S(t | H_2))
\end{equation}
This is unconstrained optimization problem, that can be solved by setting derivative w.r.t $t$ to zero, Thus the following equation can be obtained
\begin{equation*}
\frac{\mathcal{N}(t, \mu_1, \sigma_1^2)}{\mathcal{N}(t, \mu_2, \sigma_2^2)} = \beta
\end{equation*}
This equation reduces to the quadratic equation by taking a logarithm.
In the case $\sigma_1 \neq \sigma_2$, the solution is
\begin{equation}
\label{roots}
{t}_{1,2} = \frac{\sigma_1^2 \mu_2 - \sigma_2^2 \mu_1}{\sigma_1^2 - \sigma_2^2} 
\rpm
\frac{\sigma_1 \sigma_2 \sqrt{(\mu_1 - \mu_2)^2 + \Delta (\sigma_1^2 - \sigma_2^2)}}{\sigma_1^2 - \sigma_2^2}
\end{equation}
where $\Delta = 2 \log{\left( \beta {\sigma_1}/{\sigma_2}  \right)}$.
When $\sigma_1 > \sigma_2$, the right root gives the minimum and when $\sigma_1 < \sigma_2$, the left root is the solution of the problem \eqref{mindcf}. In the case of $\sigma_1 = \sigma_2 = \sigma$ and $\mu_1 > \mu_2$ the solution is
\begin{equation}
\label{oneroot}
t = \frac{\mu_1 + \mu_2}{2} + \frac{\sigma^2 \Delta}{2(\mu_1 - \mu_2)}
\end{equation}
The case when $\mu_1 < \mu_2$ is unusual for speaker verification task.
For speaker dependent PLDA scores, parameters $\mu_1, \sigma_1, \mu_2, \sigma_2$ depend on the speaker's enrolment, on the number of utterances in the enrolment. If these parameters for the specific speaker are known, then the value of the optimal threshold $t$ is determined using \eqref{roots}, \eqref{oneroot}. The score normalization expression that we suggest shifts minDCF point to the zero and is is following
\begin{equation*}
\label{normalization}
s_{norm} = \frac{1}{\sqrt{\sigma_1^2 + \sigma_2^2}}(s - t)
\end{equation*}
Here the empirical normalization by the total variance is applied. While shifting by speaker dependent $t$ aligns minDCF threshold for all speakers, presented scale normalization aligns scores in the vicinity of a minDCF point. Also this technique showed good results in our experiments on NIST SRE 2014 database. 
\section{Speaker dependent distribution of PLDA scores}
\label{sec:scorenorm}
In this section we derive speaker specific parameters $\mu,\sigma$ for score distributions of target and non-target hypotheses. First, standard PLDA model will be described and then score in multiple enrollment case will be presented. We approximate score distribution using Gaussian distribution and examine two cases: when test vector correlates with speaker's enrollment set, i.e. target hypothesis $H_1$, and when there is no correlation, i.e. non-target hypothesis $H_2$.

Given a speaker and a set of $i$-vectors $i_1, . . . , i_L$ , PLDA assumes \cite{prince2007probabilistic} that the $i$-vectors are distributed according to
\begin{eqnarray*}
&& i_n = m + Fx + Gy_n + e_n
\\
&& x \thicksim \mathcal{N}(x, 0, I_f), \quad y_n \thicksim \mathcal{N}(y_n, 0, I_g)
\\
&& e_n \thicksim \mathcal{N}(e_n, 0, \Sigma)
\end{eqnarray*}
where $m$ is the mean vector, $x \in \mathds{R}^f$ is a speaker factor that supposed to be the same for all $i$-vectors of the speaker, $y_n \in \mathds{R}^g$ is a channel factor, $e_n$ is a residual, $I$ is an identity matrix of respectable dimension, and $\Sigma$ is a diagonal covariance matrix. Further, for simplicity of calculations we assume that the mean vector $m$ is equal to zero. This can be achieved by preliminary subtracting it from the data.

Consider the trial containing the speaker's enrolment set $i_1, . . . , i_L$ and the test $i$-vector $\bm{i_t}$.
PLDA LLR verification score $s$ for this trial can be written \cite{larcher2013phonetically, rajan2014single} as
\begin{equation}
\label{initscore}
s = \frac{1}{2} \left[ (i + \bm{i_t})^T K_{L+1} (i + \bm{i_t}) - i^T K_{L} i - \bm{i_t}^T K_{1} \bm{i_t}  \right] + \alpha(L)
\end{equation}
where $i = \sum_{n=1}^L i_n$ - the sum of the speaker's $i$-vectors and $K_L$, $\alpha(L)$ are defined as follows
\begin{equation*}
K_L = \bar{U} F M_L^{-1} F^T \bar{U}
\end{equation*}
\begin{equation*}
\alpha(L) = \log{    \frac{\det \left(M_{L+1}\right)^{-1}} 
{\det \left(M_{L}\right)^{-1}
\cdot
\det \left(M_{1}\right)^{-1}} }
\end{equation*}
where $U=GG^T + \Sigma, \quad \bar{U} = U^{-1}, \quad M_L = L \cdot F^T \bar{U} F + I_f$. Further, to derive speaker dependent distribution we assume that the speaker's enrollment is known and $\bm{i_t}$ is random variable.
To deduce this distribution, the expression for the score \eqref{initscore} is rewritten as a quadratic form in the variable $\bm{i_t}$
\begin{equation}
\label{quadraticplda}
s = \frac{1}{2} (\bm{i_t} - d)^T A (\bm{i_t} - d)^T + c - \frac{1}{2} b^T A^{-1}b
\end{equation}
using following notations
\begin{eqnarray*}
 A = K_{L+1} - K_{1}, \quad && b = K_{L+1}i
\\
c = \frac{1}{2} i^T (K_{L+1} - K_{L}) i  + \alpha(L), \quad && d = - A^{-1} b
\end{eqnarray*}
Parameters of the quadratic form depend on the enrollment size $L$ and the sum of enrollment $i$-vectors $i$.
Here should be considered distribution of the quadratic form with normal distributed vector $\bm{i_t}$. There has been some works on computing such distributions \cite{castano2005distribution}, \cite{liu2009new}, \cite{bausch2013efficient}, \cite{kuonen1999miscellanea}, \cite{sheil1977algorithm}.
Quadratic form can be rewritten as a linear combination of independent non-central chi-squared distributed variables by using the transition to the new variables associated with the principal components of the matrix $A$. The convolution of this distributions leads to the complex distribution for which the solution of the problem \eqref{mindcf} seems difficult. In the $\bm{i}$-vector case \eqref{quadraticplda}, this sum consists of 400 to 600 elements since it is the typical dimension of $\bm{i}$-vector space. In this work, we approximate this distribution by using Gaussian distribution and reserve the case of non-Gaussian distribution for possible further research.

Consider quadratic form $z = q^T\Lambda q$, where $q$ is the random vector with the expected value $\mu_q$ and covariance matrix $\Sigma_q$. Then, the expectation and variance of $z$ are defined as follows
\begin{eqnarray}
\label{quad_expect}
&& \mu_z = \tr\left( \Lambda \Sigma_q \right) + \mu_q^T \Lambda \mu_q
\\
\label{quad_cov}
&& \sigma_z^2 = 2\tr\left( \Lambda \Sigma_q \Lambda \Sigma_q  \right) + 4 \mu_q^T \Lambda \Sigma_q \Lambda \mu_q
\end{eqnarray}
Consider first case of the non-target hypotheses $H_2$ when the test $i$-vector has zero expectation and covariance matrix $R = V + U$, where $V=FF^T$. Parameters of the distribution \eqref{init_nontar} are obtained by applying \eqref{quad_expect}, \eqref{quad_cov} to the quadratic form  \eqref{quadraticplda}
\begin{eqnarray*}
&& \mu_2 = \frac{1}{2} \tr\left( A R \right) + c
\\
&& \sigma_2^2 = \frac{1}{2} \tr\left( A R A R  \right) + b^T R b
\end{eqnarray*}
In the case of the target hypotheses $H_1$, test $i$-vector is correlated with speaker's enrolment and has more complex distribution. Combined vector $[\bm{i_t}, i_1, \hdots, i_N]^T$ has zero expectation and covariance matrix
\begin{equation*}
C = 
\begin{bmatrix} 
    U+V & V   & \hdots & V\\
    V   & U+V & \hdots & V\\
    \vdots & \vdots & \ddots & \vdots\\
    V   & V   & \hdots & U+V
\end{bmatrix} 
\end{equation*}
It can be shown \cite{rajan2014single} that conditional probability distribution of $\bm{i_t}$ given $i_1, \hdots, i_N$ has the following expectation and covariance
\begin{eqnarray*}
&& \hat{\mu} = \left( V \bar{U} + L \cdot V Q \right)i
\\
&& \hat{R} = R - L \cdot \left( V \bar{U} + L \cdot V Q \right) V
\end{eqnarray*}
where $Q = -(L \cdot V + U)^{-1} V \bar{U}$. Now, using \eqref{quad_expect} and \eqref{quad_cov}, final expressions for $\mu_1$ and $\sigma_1^2$ from \eqref{init_tar} are derived
\begin{eqnarray*}
&& {\mu}_{1} = \frac{1}{2} \tr(A \hat{R}) - \hat{\mu}^T A d +  \frac{1}{2} \hat{\mu}^T A \hat{\mu} + c
\\
&& \sigma_1^2 = \frac{1}{2}  \tr\left( A \hat{R} A \hat{R}  \right) + (d-\hat{\mu})^T A \hat{R} A (d-\hat{\mu})
\end{eqnarray*}
As a result, it is clear that the parameters of speaker dependent score distributions $\mu_1, \sigma_1, \mu_2, \sigma_2$ depend on the speaker's enrollment size $L$ and on the sum of enrollment $i$-vectors.

\section{Experimental results}
\label{sec:experiments}
\subsection{Data set and PLDA parameters estimation}
\label{ssec:dataset}
\textit{NIST $i$-vector Machine Learning Challenge 2014} data set has been chosen to test the efficiency of the proposed model. The data set consists of a labeled development set (\textit{devset}), a labeled model set (\textit{modelset}) with 5 $i$-vectors per model and an unlabeled test set (\textit{testset}). Since labels for the \textit{devset} were not available during the challenge, the best results were obtained from methods that allowed to cluster the  \textit{devset} and then to apply PLDA \cite{khoury2014hierarchical, novoselov2014stc}. The original \textit{devset} labels have been used in the presented experiments.

In our experiments the datasets have been initially preprocessed. Preliminary all $i$-vectors with duration less then 10 seconds have been removed \cite{khoury2014hierarchical, novoselov2014stc}. We construct a new labeled \textit{trainset}, \textit{modelset}, \textit{testset}, \textit{modelsetCV}, \textit{testsetCV}.
Speakers from \textit{devset} with 3 to 10 $i$-vectors combined with the initial \textit{modelset} are assigned to the \textit{trainset}, with 11 to 15 $i$-vectors are assigned to the new \textit{modelset} and \textit{testset}, remaining speakers with more then 15 $i$-vectors form cross validation set (\textit{modelsetCV}, \textit{testsetCV}). First 5 $i$-vectors from each speaker set are used as enrollment in the \textit{modelset} and the remaining as the \textit{testset}. The same is done for the cross validation set. Eventually the \textit{trainset} contains 3281 speakers and total 18759 $i$-vectors, 717 speakers with 3585 $i$-vectors and  5400 $i$-vectors in the \textit{modelset} and the \textit{testset} respectively.We used minDCF with $\beta = 100$ as a measure of the system performance.

The whitening process is applied to the \textit{trainset} \cite{garcia2011analysis}. Whitened \textit{trainset} is used for the PLDA model parameter estimation. The parameters of whitening are computed on the \textit{trainset} too. Whitened \textit{trainset} was projected on the unit sphere \cite{garcia2011analysis}. This transform is used further for all trials. Best speaker and channel factor dimensions for PLDA are equal to 590 and 10 respectively. Optimal parameters were found on cross validation set - \textit{modelsetCV}, \textit{testsetCV}.
\subsection{Results}
\label{ssec:results}
The experiments were performed with various enrollment size conditions. \textit{modelset} contains speakers with 5 $i$-vectors per speaker. We compare general PLDA scoring with suggested normalized scoring \eqref{normalization} on the following enrollment sets. Fist, 5 sets with $L=\{1,2,3,4,5\}$ $i$-vectors per speaker are configured using \textit{modelset}. For example, the set with $L=3$ is composed using first 3 $i$-vectors of each speaker from \textit{modelset}. In addition, the set with the mixed enrollment conditions is configured. This set is composed from \textit{modelset} speakers with the reduced enrollment and contain 94 speakers with 1 $i$-vector per speaker, 93 speakers with 2 $i$-vectors per speaker, 194 speakers with 3 $i$-vectors per speaker, 189 speakers with 4 $i$-vectors per speaker, 113 speakers with 5 $i$-vectors per speaker.

\begin{figure}[htb]

\begin{minipage}[b]{1.0\linewidth}
  \centering
  \centerline{\includegraphics[width=9.0cm]{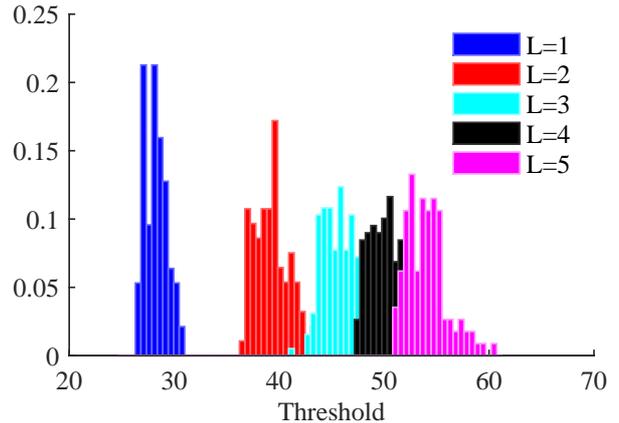}}
\end{minipage}

\caption{Histograms of speakers thresholds with various enrollment conditions}\medskip
\label{fig:ThreshHist}
\end{figure}

Figure~\ref{fig:ThreshHist}. demonstrates the histograms of the speaker dependent thresholds for models with various enrollment sizes $L$. These thresholds are found to be optimal in terms of minDCF \eqref{mindcf} with $\beta = 100$. As could be seen, the means of threshold distribution differ based on size of enrollment set, and variances indicates uncertainties of different speaker scores.
\begin{table}[h]
    \begin{center}
        \begin{tabular}{  c  c  c  c }
        \hline
                        & $L=1$  & $L=2$  & $L=3$\\ \hline
        PLDA            & 0.361  & 0.209  & 0.159\\
        Norm. PLDA      & 0.362  & 0.210  & 0.159\\
        \\
        \hline
                        & $L=4$  & $L=5$  & Mix. enroll.\\ \hline
        PLDA            & 0.131  & 0.113  & 0.266\\ 
        Norm. PLDA      & 0.131  & 0.116  & \textbf{0.188}\\
        \end{tabular}
        \caption{Results on the NIST SRE 2014 with various enrollment conditions (minDCF with $\beta = 100$).}
        \label{table1}
    \end{center}
\end{table}
Table~\ref{table1} demonstrates results for all sets. In the case of the uniform enrollment conditions $L=\{1,2,3,4,5\}$, the suggested normalization technique shows almost the same performance as the standard PLDA scoring. Much better results are achieved on the set with mixed enrollment conditions. As it is expected, using constant threshold gives worse results than by using proposed normalization. At the same time it doesn't bring extra computational costs since parameter estimation is made in a blind manner. This experiments shows that the performance really degrades with unfixed enrollment size and could be enhanced with this cheap procedure. In the current challenge normalized scoring decreases minDCF by 30\% comparing with general PLDA scoring.
\section{Conclusions and Further Work}
\label{sec:conclusions}
This paper presents a novel normalization technique for $i$-vector PLDA speaker verification 
in the mixed enrollment number condition. The main contribution to existed normalization methods is that this technique does not require extra development data and based only on the properties of the PLDA model. This provide more stable verification scores almost without additional computational costs.  The experiments conducted on NIST SRE 2014 database demonstrate that minDCF decrease in the mixed enrollment number condition.

In further work, the problem of non-Gaussian distribution for PLDA scores can be considered. This can lead to a more accurate estimate of the speaker specific threshold.
\section{Acknowledgement}
\label{sec:acknowledgement}
Research is conducted by Stel - Computer systems ltd. with support of the Ministry of Education and Science of the Russian Federation (Contract №14.579.21.0058)
Unique ID for Applied Scientific Research (project) RFMEFI57914X0058.
The data presented, the statements made, and the views expressed are solely the responsibility of the authors.

\newpage
\eightpt

\bibliographystyle{IEEEtran}
\bibliography{BlindPLDAinter.bib}

\end{document}